\documentclass[letterpaper]{article} % DO NOT CHANGE THIS
\usepackage{aaai25}  % DO NOT CHANGE THIS
\usepackage{times}  % DO NOT CHANGE THIS
\usepackage{helvet}  % DO NOT CHANGE THIS
\usepackage{courier}  % DO NOT CHANGE THIS
\usepackage[hyphens]{url}  % DO NOT CHANGE THIS
\usepackage{graphicx} % DO NOT CHANGE THIS
\usepackage{natbib}  % DO NOT CHANGE THIS AND DO NOT ADD ANY OPTIONS TO IT
\usepackage{caption} % DO NOT CHANGE THIS AND DO NOT ADD ANY OPTIONS TO IT
\usepackage{booktabs}
\usepackage{multirow}
\usepackage{array}
\usepackage{amsmath}
\usepackage{amssymb}
\usepackage{amsthm}
\usepackage{graphicx}
\usepackage{algorithm}
\usepackage{algorithmic, mathrsfs, bm}
\usepackage{appendix}
\usepackage{footmisc}
\usepackage{colortbl}
\usepackage{cleveref}
\newtheorem{theorem}{Theorem}
\newtheorem{lemma}{Lemma}
\usepackage{subcaption}
\newtheorem{assumption}{Assumption}

\usepackage{newfloat}
\usepackage{listings}
\DeclareCaptionStyle{ruled}{labelfont=normalfont,labelsep=colon,strut=off} % DO NOT CHANGE THIS
\floatstyle{ruled}
\newfloat{listing}{tb}{lst}{}
\floatname{listing}{Listing}
\title{Asymptotic Unbiased Sample Sampling to Speed Up Sharpness-Aware Minimization}
\author {
    Jiaxin Deng\textsuperscript{\rm 1},
    Junbiao Pang\textsuperscript{\rm 1}\thanks{Corresponding authors.},
    Baochang Zhang\textsuperscript{\rm 2,\rm 3},
    Guodong Guo\textsuperscript{\rm 4}
}
\affiliations {
    \textsuperscript{\rm 1}School of Information Science and Technology, Beijing University of Technology, Beijing, China\\
    \textsuperscript{\rm 2}Hangzhou Research Institute, School of Artificial Intelligence, Beihang University, China\\
    \textsuperscript{\rm 3}Nanchang Institute of Technology, Nanchang, China\\
    \textsuperscript{\rm 4}Ningbo Institute of Digital Twin, Eastern Institute of Technology, Ningbo, China\\
    dengjiaxin@emails.bjut.edu.cn, junbiao\_pang@bjut.edu.cn, bczhang@buaa.edu.cn, guodong.guo@mail.wvu.edu
}

\usepackage{bibentry}
\begin{document}

\maketitle

\begin{abstract}
Sharpness-Aware Minimization (SAM) has emerged as a promising approach for effectively reducing the generalization error.
However, SAM incurs twice the computational cost compared to the base optimizer (\textit{e.g.}, SGD).
We propose Asymptotic Unbiased data sampling to accelerate SAM (AUSAM), which maintains the model's generalization capacity while significantly enhancing computational efficiency.
Concretely, we probabilistically sample a subset of data points beneficial for SAM optimization based on a theoretically guaranteed criterion, \textit{i.e.}, the Gradient Norm of each Sample (GNS).
We further approximate the GNS by evaluating the difference in loss values before and after perturbation in SAM. As a plug-and-play, architecture-agnostic method, our approach consistently accelerates SAM across various tasks and networks, \textit{i.e.,} classification, human pose estimation, and network quantization.
On CIFAR-10/100 and Tiny-ImageNet, AUSAM achieves results comparable to SAM while providing a speedup of over 70\%. By adjusting hyperparameters, AUSAM can match the speed of the base optimizer while significantly surpassing the base optimizer's performance. 
Compared to recent dynamic data pruning methods, AUSAM is better suited for SAM and excels in maintaining performance. Additionally, AUSAM accelerates optimization in human pose estimation and model quantization without sacrificing performance, demonstrating its broad practicality.
\end{abstract}

\section{Introduction}

The generalization capacity of a model significantly impacts its effectiveness and reliability in practical applications, making it a consistently important topic in deep learning research \cite{neyshabur2017exploring,izmailov-2018-SWA-UAI,cha2021swad,andriushchenko-2022-towards_understanding-ICML,zhang2023flatness}.
Several studies have delved into the correlation between generalization performance and the geometry of the loss landscape~\cite{keskar-2016-large_batch-ICLR,jiang-2019-fantastic-ICLR,wu2020adversarial,zhang-2021-understanding-ACM}. 
The loss landscape is complex and non-convex, with many local minima that exhibit varying generalization performance~\cite{hochreiter-1994-simplifying-NIPS}. These studies observed that flatter minima on the loss landscape tend to generalize more effectively than sharper minima. This motivates the idea of finding flatter minima during training.

Foret et al. recently introduced the Sharpness-Aware Minimization (SAM) \cite{foret-2020-SAM-ICLR} algorithm, which offers an elegant approach to seek flat minima and improve generalization.
The SAM prevents the model from converging to sharp minima, thereby enhancing generalization. 
SAM and its variants have demonstrated state-of-the-art performance across various applications \cite{kwon-2021-asam-ICML,jiang-2022-aesam-ICLR,zhuang-2022-GSAM-ICLR}.
However, SAM lacks efficiency due to its requirement of two forward and two backward in each optimization step.
In recent years, various methods have been proposed to accelerate SAM. These can be roughly divided into methods for reducing gradient computations, such as G-RST \cite{zhao2022randomized}, LookSAM \cite{liu-2022-looksam-CVPR}, AE-SAM \cite{jiang-2022-aesam-ICLR}, vSAM \cite{deng2024effective} and methods for efficient SAM optimization processes, such as SAF \cite{du-2022-saf-NIPS}, K-SAM \cite{ni2022k} and ESAM \cite{du-2022-ESAM-ICLR}. 
However, most of them tends to potentially reduce the model's performance, especially the acceleration approach by reducing the number of SAM updates. Although some strategies can help models to maintain the performance, the extra operations introduced by implementing these strategies may limit gains in optimization efficiency.

In this paper, we introduce a dynamic sampling strategy for each mini-batch, based on maintaining expected sharpness across sampled and original data. Our theoretical analysis indicates that samples with larger gradient norms are crucial for preserving the model's generalization. Thus, we assign a score to each sample during forward propagation according to its gradient norm and prioritize higher-scoring samples in each mini-batch for training.
This approach significantly improves optimization efficiency while preserving the model's generalization ability without excessive compromise. Compared to data pruning methods \cite{toneva2018empirical,paul2021deep,raju2021accelerating}, our method reduces the gradient expectation bias between the entire batch and the selected samples, achieving asymptotic unbiasedness as training progresses.

Our contributions are summarized as follows: 
\begin{itemize}
  \item We discovered that implementing SAM with mini-batch subsets characterized by high gradient norms effectively preserves the generalization capabilities of the model.
  
  \item We proposed an asymptotically unbiased data sampling strategy that accelerates the SAM optimization process without significantly compromising the model's generalization ability.
  
\end{itemize}

\section{Related Work}
\subsection{Sharpness-Aware Minimization}
Foret et al. introduced SAM~\cite{foret-2020-SAM-ICLR}, which aims to enhance the generalization ability of the model.
SAM can be viewed as addressing a minimax optimization problem as follows:
\begin{equation}\label{equ:sam}
\begin{aligned}
\mathop {\min } \limits_\mathbf{w}  \; &{L^{SAM}}(\mathbf{w}) + \lambda ||\mathbf{w}||^2  \\ &\text{where} \; {L^{SAM}}(\mathbf{w}) = \mathop {\max }\limits_{||\bm{\varepsilon} || \le \rho } L(\mathbf{w} + \bm{\varepsilon} ),
\end{aligned}
\end{equation}
where $\bm{\varepsilon}$ represents weight perturbations in an Euclidean ball with the radius $\rho$, $\lambda ||\mathbf{w}||^2$ is the standard L2 regularization term. SAM utilizes Taylor expansion to search for the maximum perturbed loss in local parameter space:
\begin{equation}\label{equ:psf}
\begin{aligned}
\mathop {\arg \max }\limits_{||\bm{\varepsilon} || \le \rho } \; L(\mathbf{w} + \bm{\varepsilon} )  & \approx \mathop {\arg \max }\limits_{||\bm{\varepsilon} || \le \rho } \; L(\mathbf{w}) + {\bm{\varepsilon} ^T}{\nabla _\mathbf{w}}L(\mathbf{w}) \\
&= \mathop {\arg \max }\limits_{||\bm{\varepsilon} || \le \rho } \; {\bm{\varepsilon} ^T}{\nabla _\mathbf{w}}L(\mathbf{w}).
\end{aligned}
\end{equation}
By solving Eq.~\eqref{equ:psf}, SAM can obtain the perturbation $\hat{ \bm{\varepsilon} } = \rho {\nabla _\mathbf{w}}L(\mathbf{w})/||{\nabla _\mathbf{w}}L(\mathbf{w})||$ that maximizes the loss function $L^{SAM}(\mathbf{w})$.
By substituting the perturbation back into Eq.~\eqref{equ:sam} and the gradient approximation, the gradient during the weight optimization process in SAM is described as follows:
\begin{equation}\label{equ:grad_sam}
\begin{aligned}
{\nabla _\mathbf{w}}{L^{SAM}}(\mathbf{w}) \approx {\nabla _\mathbf{w}}L(\mathbf{w} + \bm{\hat\varepsilon} ) \approx {\nabla _\mathbf{w}}L(\mathbf{w}){|_{\mathbf{w} + \bm{\hat\varepsilon} }}.
\end{aligned}
\end{equation}
From Eq.~\eqref{equ:psf} and Eq.~\eqref{equ:grad_sam}, we observe that SAM necessitates two forward and two backward  to update weights once.
This makes the computational overhead of SAM twice that of the basic optimizer.

\subsection{Methods of Accelerating SAM}
In this section, we discuss related work on methods for reducing gradient computations, methods for efficient SAM optimization processes, and dynamic data pruning methods.

\textbf{Methods for reducing gradient computations.} The motivation behind these methods is to reduce the number of times SAM optimization is utilized during the training.
G-RST~\cite{zhao2022randomized} performs a Bernoulli trial at each iteration, randomly selecting between SGD and SAM based on probabilities determined by a predefined scheduling function. 
LookSAM~\cite{liu-2022-looksam-CVPR} employs SAM at every $k$ iterations and reuses previous gradients in other iterations.
AE-SAM~\cite{jiang-2022-aesam-ICLR} adaptively employs SAM based on the loss landscape geometry.
However, these methods often require extra operations to achieve optimization acceleration, which limits the overall speedup.

\textbf{Methods for efficient SAM optimization processes.} These methods aim to improve the efficiency of each step of SAM.
SAF and MESA~\cite{du-2022-saf-NIPS} employ the trajectory loss to approximate sharpness.
However, SAF and MESA do not employ direct weight perturbations to help SAM escape local minimum regions.
\cite{bahri2021sharpness} reveals SAM’s benefits when data is limited by fine-tuning on sub-samples of the original task training split. But it randomly selects samples.
K-SAM~\cite{ni2022k} proposes to compute gradients in both stages of SAM on only the top-k samples with highest loss. 
ESAM~\cite{du-2022-ESAM-ICLR} uses fewer samples to compute the gradients and only updates part of the model in the SAM's second step.
However, ESAM requires extra computation when selecting samples.
In Momentum-SAM~\cite{becker2024momentum}, parameters are perturbed along the direction of the accumulated momentum vector, aiming to reduce sharpness and promote flatter minima.
Although these methods improve the optimization efficiency, they may lead to a significant decrease in the generalization performance of the model. 

\textbf{Dynamic data pruning methods.}
Dynamic data pruning aims to reduce training costs by reducing the number of iterations for training \cite{toneva2018empirical,paul2021deep}. 
The pruning process is conducted during training, and sample information can be obtained during training.
Dynamically prunes the samples based on easily attainable scores, e.g., loss values, during the training without trials.
\cite{raju2021accelerating} proposes three different data selection criteria in dynamic data pruning. 
Uncertainty sampling preferentially selects the sample with the lowest confidence in the current prediction of the model. 
$\varepsilon$-greedy approach is inspired by reinforcement learning.
The upper-confidence bound (UCB) approach suggests pruning samples based on the upper confidence bound of their losses.
InfoBatch \cite{qin-2024-infobatch} randomly prunes less informative samples based on loss distribution and rescales remaining samples' gradients to approximate the original gradient. However, these methods are not specific to SAM.

\section{Method}
\subsection{Principle of Sampling}
We aim to select a subset of samples from each mini-batch during training to reduce the computational costs of SAM's forward and backward propagation.

\begin{theorem}\label{thm:diff_uperbound}
 Suppose we select $N$ samples from a mini-batch $\mathbf{B}$ of size $K$, denoting them as $\hat{\mathbf{B}}$ ($\hat{{x}_i} \in \hat{\mathbf{B}}$, $i=1,2,...,N$), and denote the remaining set of $M(M=K-N)$ samples as $\check{\mathbf{B}}$ ($\check{{x}_i} \in \check{\mathbf{B}}$, $i=1,2,...,M$). Let $\hat{\bm{\varepsilon}}_\mathbf{B}$ and $\hat{\bm{\varepsilon}}_{\hat{\mathbf{B}}}$ represent the parameter perturbations in SAM obtained through the set $\mathbf{B}$ and $\hat{\mathbf{B}}$, respectively.
 The difference in the Perturbation loss and Training loss between the calculations using a Full mini-batch (PTF) and those using Sampled samples (PTS) is as follows:
%\begin{tiny}
\begin{equation}\label{equ:diff_loss}
\begin{aligned}
&\bigg| \underbrace{\left[ {{L_\mathbf{B}}(\mathbf{w} + {{\hat {\bm{\varepsilon}} }_\mathbf{B}}) - {L_\mathbf{B}}(\mathbf{w})} \right]}_{\text{PTF}} 
- 
\underbrace{\left[ {{L_{\hat{\mathbf{B}}}}(\mathbf{w} + {{\hat {\bm{\varepsilon}} }_{\hat {\mathbf{B}}}}) - {L_{\hat {\mathbf{B}}}}(\mathbf{w})} \right]}_{\text{PTS}} \bigg| \\
&\le \frac{\rho }{M}\sum\limits_{i = 1}^M {\left\| {{\nabla _\mathbf{w}}{L_{{\check{{x}_i}}}}(\mathbf{w})} \right\|}  + \left|(O({{\hat {\bm{\varepsilon}} }_\mathbf{B}}^2) - O({{\hat {\bm{\varepsilon}} }_{\hat {\mathbf{B}}}}^2))\right|.
\end{aligned}
\end{equation} 
%\end{tiny}
\end{theorem}
When we ignore the error term, Theorem \ref{thm:diff_uperbound} demonstrates that the upper bound for the difference between PTF and PTS is smaller when the gradient norm of the unsampled samples is lower.
One possible sampling method uses the gradient norm of each sample as the selection criterion.
However, we still face a problem: how to efficiently compute the gradient norm of each sample $\left\| {{\nabla _\mathbf{w}}{L_{\mathbf{x}_i}}(\mathbf{w})} \right\|$ as the criterion?

\subsection{Efficiently Per-sample gradient vs. Directly Approximate Gradient Norms}

\textbf{Efficiently computing per-sample gradient norm.}
The naive approach to computing per-sample gradient utilizes back-propagation when the mini-batch size is 1. However, this approach has significantly low optimization efficiency.~\cite{goodfellow2015efficient}~\cite{rochette2019efficient} used the auto differentiation’s intermediate results to compute the per-example gradient. However, these methods still require the extra operations to obtain the per-sample gradient. For instance, \cite{goodfellow2015efficient} has a time complexity of $O(Kmp^2)$ to compute the gradient of $K$ samples, where $m$ is the number of the layers in a neural network, and $p$ is the number of dimensions in each layer. An extra time complexity of $O(K)$ is also incurred to compute gradient norms. A natural question is: is it possible to directly approximate per-sample gradient norms from the results of the perturbation in Eq.~\eqref{equ:psf}?

\begin{lemma}\label{lemma:lemma1}
For a mini-batch $\mathbf{B}$, the gradient at 
$\mathbf{w}$ is obtained as  ${{\nabla _\mathbf{w}}{L_\mathbf{B}}(\mathbf{w})}$ and the parameter perturbation in SAM is ${\bm{\varepsilon} _\mathbf{B}}$.
The training loss and perturbation loss of sample $\mathbf{x}_i \in \mathbf{B}$ denote as ${L_{\mathbf{x}_i}}(\mathbf{w})$ and ${L_{\mathbf{x}_i}}(\mathbf{w} + {\bm{\varepsilon} _\mathbf{B}})$ respectively. When the gradient ${{\nabla _\mathbf{w}}{L_\mathbf{B}}(\mathbf{w})}$ is nonzero, the gradient norm of sample $\mathbf{x}_i$ is approximated as follows:
\begin{equation}\label{equ:diff_loss}
\begin{aligned}
&\left\| {{\nabla _\mathbf{w}}{L_{{\mathbf{x}_i}}}(\mathbf{w})} \right\| \\
&\approx \left\| {\frac{{{L_{{\mathbf{x}_i}}}(\mathbf{w} + \rho \frac{{{\bigtriangledown_\mathbf{w}}{L_\mathbf{B}}(\mathbf{w})}}{{|| {{\bigtriangledown_\mathbf{w}}{L_\mathbf{B}}(\mathbf{w})}||}}) - {L_{{\mathbf{x}_i}}}(\mathbf{w})}}{\rho }\frac{{{\bigtriangledown_\mathbf{w}}{L_\mathbf{B}}(\mathbf{w})}}{{|| {{\bigtriangledown_\mathbf{w}}{L_\mathbf{B}}(\mathbf{w})} ||}}} \right\| \\ 
&= \frac{1}{\rho } \underbrace{|{{L_{{\mathbf{x}_i}}}(\mathbf{w} + {\bm{\varepsilon} _\mathbf{B}}) - {L_{{\mathbf{x}_i}}}(\mathbf{w})|}}_{DLP} .
\end{aligned}
\end{equation}
\end{lemma}

\textbf{Directly approximate gradient norm.}\label{sec:estimate-average-gradient}
Lemma~\ref{lemma:lemma1} shows that the gradient norm of sample $\mathbf{x}_i$ can be approximated by the absolute values of \textbf{D}ifference in \textbf{L}oss before and after \textbf{P}erturbation (hereinafter referred to as DLP), \textit{i.e.},  ${|{L_{\mathbf{x}_i}}(\mathbf{w} + {\bm{\varepsilon}_{\mathbf{B}}}) - {L_{\mathbf{x}_i}}(\mathbf{w})}|$. Therefore, instead of estimating the gradient norm $\left\| {{\nabla _\mathbf{w}}{L_{\mathbf{x}_i}}(\mathbf{w})} \right\|$, we use DLP as our sampling indicator. To improve the accuracy of the gradient norm evaluation, we estimate the current gradient norm by dynamically averaging the historical DLP for each sample. The Average DLP (ADLP) for sample $\mathbf{x}_i$ at epoch $T$ is given by the following:
\begin{equation}\label{equ:avg}
\begin{aligned}
g_{{\mathbf{x}_i}}^T = \frac{1}{{T - 1}}\sum\limits_{t = 1}^{T - 1}{|{L_{{\mathbf{x}_i}}}(\mathbf{w}_t + {\bm{\varepsilon}_{\mathbf{B}_t}}) - {L_{{\mathbf{x}_i}}}(\mathbf{w}_t)|}.
\end{aligned}
\end{equation}
The time and storage complexity of Eq.~\eqref{equ:avg} are $O(1)$ and $O(D)$, respectively, where $D$ is the number of samples in the dataset.

\subsection{Sampling Strategy} \label{sec:sampling-strategy}
To ensure that samples with low $g_{{\mathbf{x}_i}}^T$ values are not overlooked, we implement a probabilistic sampling mechanism. The sampling probability for each sample at epoch $T$ is determined as follows:
\begin{equation}\label{equ:pxi}
\begin{aligned}
{p^T_{\mathbf{x}_i}} = \frac{{{{g}^T_{{\mathbf{x}_i}}}}}{{\sum\nolimits_{i = 1}^K {{{g}^T_{{\mathbf{x}_i}}}} }},
\end{aligned}
\end{equation}
where $p^T_{\mathbf{x}_i}$ denotes the sampling probability of sample $\mathbf{x}_i$ at epoch $T$. To control the ratio of the maximum and minimum sampling probabilities, we normalize ${g}^T_{{\mathbf{x}_i}}$ using min-max normalization as follows:
\begin{equation}\label{equ:dxi}
\begin{aligned}
\tilde{g}^T_{{\mathbf{x}_i}} = {s_{min}} + \frac{{{g}^T_{{\mathbf{x}_i}} - {g}_{\mathbf{B}}^{min}}}{{{g}_{\mathbf{B}}^{max} - {g}_{\mathbf{B}}^{min}}}({s_{max}} - {s_{min}}),
\end{aligned}
\end{equation}
where $s_{min}$ and $s_{max}$ represent the lower and upper bounds of the normalization, respectively. ${g}_{\mathbf{B}}^{max}$ and ${g}_{\mathbf{B}}^{min}$ are the maximum and minimum values of ${g}^T_{{\mathbf{x}_i}}$ within the mini-batch $\mathbf{B}$, respectively. 
Then, we calculate the probability ${p^T_{\mathbf{x}_i}}$ by substituting ${g}^T_{{\mathbf{x}_i}}$ with $\tilde{g}^T_{{\mathbf{x}_i}}$ in Eq.~\eqref{equ:pxi}. In addition, we use a hyperparameter $\alpha$ to represent the proportion of samples selected from a mini-batch of size $K$, where $\alpha K$ samples are sampled. The time and storage complexity of this process are $O(K)$ and $O(1)$, respectively.

We gradually select samples with large gradient norms for stable initialization of training.
We define a fixed epoch $E_{start}$ and allow $s_{max}$ to be a floating value before epoch $E_{start}$, linearly increasing from $s_{min}$ to $s_{max}$.
This enables samples with larger gradient norms to be increasingly prioritized for selection during training. Algorithm \ref{algorithm:1} shows the overall proposed algorithm.

\subsection{Speed up ratio of SAM}
Suppose that the time complexity of optimizing the model with SAM is $O(T)$. We can roughly divide the total time $T$ into two parts: the forward and backward time $T_f$ and the remaining time $T_b$. Our method can theoretically reduce the forward and backward time to $\alpha T_f$. When $T_b$ is sufficiently small, our method achieves nearly $O(0.5T)$ for $\alpha=0.5$, which is almost as fast as the base optimizer.

\subsection{Theoretical Analysis}

\begin{assumption}\label{assumpion:smooth}
(Smoothness). $L(\mathbf{w})$ is $\tau$-Lipschitz smooth in $\mathbf{w}$, i.e., $\left\| {\nabla L(\mathbf{w}) - \nabla L(\mathbf{v})} \right\| \le \tau \left\| {\mathbf{w} - \mathbf{v}} \right\|$.
\end{assumption}
\begin{assumption}\label{assumpion:Bounded_g}
(Bounded gradients). By the assumption that an upper bound is exists on the gradient of sampled set $\hat{\mathbf{B}}$ from each mini-batch. There exists $G > 0$ for $\hat{\mathbf{B}}$ such that $\mathbb{E}\left[ {\left\| {\nabla L_{\hat{\mathbf{B}}}(\mathbf{w})} \right\|} \right] \le G$.
\end{assumption}
\begin{assumption} \label{assumpion:bound_var}
(Bounded variance of stochastic gradients). Given the training set $\mathbf{D}$ and a sampled set $\hat{\mathbf{B}} \in \mathbf{D}$. There exists $\sigma \ge 0$, the variance of the sampled set of size $\alpha K$ is bounded by $\mathbb{E}\left[ {{{\left\| {\nabla {L_{\hat{\mathbf{B}}}}(\mathbf{w}) - \nabla {L_\mathbf{D}}(\mathbf{w})} \right\|}^2}} \right] \le \frac{\sigma ^2}{\alpha K}$.
\end{assumption}

\begin{theorem}\label{thm:deviation}
Suppose Assumption \ref{assumpion:smooth} and \ref{assumpion:Bounded_g} hold. Let learning rate ${\eta _t} = {{{\eta _0}}}/{{t^2}}$, ${\eta _0}$ represents the initial learning rate, and there are $N$ iterations in one epoch. The deviation between the average gradient norm of sample $\mathbf{x}_i$ over $T-1$ epochs and the gradient norm at the $T$-th epoch is bounded within a specific range, defined as follows:
\begin{equation}\label{equ:deviation}
\begin{aligned}
\left| {\left\| {\nabla {L_{{\mathbf{x}_i}}}({\mathbf{w}_T})} \right\| - \frac{1}{{T - 1}}\sum\limits_{t = 1}^{T - 1} {\left\| {{\nabla _\mathbf{w}}{L_{{\mathbf{x}_i}}}({\mathbf{w}_t})} \right\|} } \right|
\le \frac{{\tau {\eta _0} \pi^2 NG}}{6}.
\end{aligned}
\end{equation}
\end{theorem}
From Theorem \ref{thm:deviation}, it is evident that the error is limited to a certain range. This suggests the viability of substituting the current gradient norm with the average gradient norm (Eq.~\ref{equ:avg}).

\begin{algorithm}[tb]
    %\begin{small}
    \caption{{Pseudocode of the proposed method}}
    {\bf Require:}
    The training dataset $\mathbf{D}$, the learning rate $\eta$, the batch size $K$, parameters $\alpha$, $\rho$, $s_{min}$ and $s_{max}$.
    \begin{algorithmic}[1]
    \FOR{$t = 1,2,\cdot\cdot\cdot$}
    \STATE Randomly sample a mini-batch $\mathbf{B}$ with size K;
    \STATE Calculate the probability for each sample in set $\mathbf{B}$ of being selected according to Eq.~\eqref{equ:pxi}.
    \STATE Sampling $\alpha K$ data from $\mathbf{B}$ to constitute a subset $\hat{\mathbf{B}}$;
    \STATE Calculate the loss for each selected sample $L_{\mathbf{x}_i}(\mathbf{w}_t)$ and gradient of all selected samples $\nabla{L_{\hat{\mathbf{B}}}}(\mathbf{w}_t)$.
    \STATE Calculate perturbation $\hat{ \bm{\varepsilon} } = \rho \frac{\nabla L_{\hat{\mathbf{B}}}(\mathbf{w}_t)}{||{\nabla }L_{\hat{\mathbf{B}}}(\mathbf{w}_t)||}$ and gradient after perturbation $\nabla {L_{\hat{\mathbf{B}}}}(\mathbf{w}_t+\bm{\varepsilon})$.
    \STATE Update the weights by ${\mathbf{w}_{t+1}} = {\mathbf{w}_{t}} - \eta \nabla {L_{\hat{\mathbf{B}}}}(\mathbf{w}_t+\bm{\varepsilon})$;
    \STATE Update the ADLP for each selected sample based on Eq.~\eqref{equ:avg}.
    \ENDFOR
    \end{algorithmic}
    \label{algorithm:1}
   % \end{small}
\end{algorithm}

\begin{theorem}\label{theorem3}
Suppose Assumptions \ref{assumpion:smooth} and \ref{assumpion:bound_var} hold. Let $K$ be the mini-batch size, $\alpha$ is the sampling rate. For learn rate $\eta  \le \frac{1}{\tau }$ and $\rho  \le \frac{1}{{4\tau }}$, the algorithm satisfies:
\begin{equation}\label{equ:ca}
\begin{aligned}
&\mathbb{E} \left [ \sum\limits_{t = 1}^T {{{\left\| \nabla L_{\mathbf{D}}({\mathbf{w}_t}) \right\|}^2}}\right ] \\ 
& \le \frac{{16}}{{7\eta T}}\left( {L[({\mathbf{w}_0})] - E[L({\mathbf{w}_T})]} \right) + \frac{{17\eta \tau {\sigma ^2}}}{{7\alpha K}}.
\end{aligned}
\end{equation}
\end{theorem}
From Theorem \ref{theorem3}, it has been observed that convergence in sample selection for training remains feasible, with tighter bounds resulting from selecting more samples from each mini-batch.

\begin{theorem}\label{thm:unbiased}
Assuming sample $\mathbf{x}$ from $\mathbf{D}$ are drawn from continuous distribution $q(\mathbf{x})$. $p^t_\mathbf{x}$ is the probability that sample $\mathbf{x}$ is sampled in the $t$-th epoch.
The perturbation of SAM is assumed to be the theoretical perturbation over the entire dataset $\mathbf{D}$, i.e. ${\bm{\varepsilon}_\mathbf{D}} = \rho {{{\nabla _\mathbf{w}}{L_D}({\mathbf{w}_t})}}//{{\left\| {{\nabla _\mathbf{w}}{L_D}({\mathbf{w}_t})} \right\|}}$. 
The expected difference between the perturbation loss and the training loss on dataset $\mathbf{D}$ and its sampled subsets $\hat{\mathbf{D}}$ can be expressed as:
\end{theorem}
% discuss why unbasis method is unnecessay 
\begin{equation}\label{equ:biased}
\begin{aligned}
&\bigg | \mathop {\mathbb{E}}\limits_{\mathbf{x} \in \mathbf{D}} \left[ {\mathop {\max }\limits_{\left\| {{\bm{\varepsilon} _\mathbf{D}}} \right\| \le \rho } [{L_\mathbf{x}}({\mathbf{w}_t} + {\bm{\varepsilon} _\mathbf{D}}) - {L_\mathbf{x}}({\mathbf{w}_t})]} \right] \\ &\quad \quad- \mathop {\mathbb{E}}\limits_{{\mathbf{x}} \in \hat{\mathbf{D}}} \left[ {\mathop {\max }\limits_{\left\| {{\bm{\varepsilon} _\mathbf{D}}} \right\| \le \rho } [{L_\mathbf{x}}({\mathbf{w}_t} + {\bm{\varepsilon} _\mathbf{D}}) - {L_\mathbf{x}}({\mathbf{w}_t})]} \right] \bigg |\\
&\le \rho \int_{\mathbf{x}} {(1 - p_\mathbf{x}^t)\left\| {{\nabla _\mathbf{w}}{L_\mathbf{x}}({\mathbf{w}_t})} \right\|q ({\mathbf{x}})} d{\mathbf{x}}.
\end{aligned}
\end{equation}

During the stage where SAM finds the internal maximum, Theorem \ref{thm:unbiased} indicates that if we choose samples with larger gradient norms $\left\| {{\nabla _\mathbf{w}}{L_{\mathbf{x}}}(\mathbf{w}_t)} \right\|$ with a higher probability $p^t_\mathbf{x}$, the expected error generated by these samples will be smaller compared to randomly selected samples. As $t$ increases, the gradient norm of sample $\mathbf{x}$ will theoretically gradually decrease, and the expected error will also gradually decrease. We still need to use these sampled samples during the SAM gradient descent phase, as they help identify the maximum loss within the parametric neighborhood space. Also, as $t$ increases, samples with smaller gradient norms become less influential. Therefore, we choose samples with larger gradient norms to better represent the full mini-batch.

\section{Experimental Results}
\label{sec:exp}

\subsection{Setup}
\label{subsec:setup}

\textbf{Datasets and Models.}
We conduct experiments on the CIFAR-10, CIFAR-100 \cite{krizhevsky2009learning} and Tiny-ImageNet \cite{le-2015-tiny} image classification benchmark datasets. 
We employ a variety of architectures, i.e., ResNet-18 \cite{he2016deep}, WideResNet-28-10 \cite{zagoruyko2016wide} and PyramidNet-110 \cite{han2017deep} on CIFAR-10 and CIFAR-100, ResNet-18, ResNet-50 \cite{he2016deep}, and MobileNetv1 \cite{howard2017mobilenets} on Tiny-ImageNet \cite{chrabaszcz2017downsampled}, to evaluate the performance and training efficiency.

\noindent\textbf{Baselines.}
We use vanilla SGD and SAM \cite{foret-2020-SAM-ICLR} as baselines.
% category these method into different groups
To comprehensively evaluate performance, we have also chosen some efficient methods ESAM \cite{du-2022-ESAM-ICLR}, LookSAM \cite{liu-2022-looksam-CVPR}, SAF \cite{du-2022-saf-NIPS}, MESA \cite{du-2022-saf-NIPS}, K-SAM \cite{ni2022k} for comparison. 
These methods are the follow-up works of SAM that aim to enhance efficiency. 
Additionally, we choose InforBatch \cite{qin-2024-infobatch}, an advanced dynamic data pruning method, and apply it to SAM. We also establish a dynamic pruning baseline, referred to as "SAM+Random", which randomly selects half of the samples from each batch.

\noindent\textbf{Implementation Details.}
We trained all the models for 200 epochs with a batch size of 128, employing cutout regularization \cite{devries2017improved} and cosine learning rate decay \cite{loshchilov2016sgdr} for all methods. For the proposed method, we set $s_{min}=0.1$, $s_{max}=0.5$ for ResNet-18 and WideResNet-28-10, $s_{max}=1$ for PyramidNet-110. Typically, $s_{max}$ is set to $1$, though for smaller networks, it may be reduced further. This adjustment is motivated by the heightened impact of omitting samples with small gradient norms on smaller networks. Lowering $s_{max}$ increases the chances for samples with smaller gradient norms to be chosen.

For Tiny-ImageNet, we also train ResNet-18, ResNet-50 and MobileNetv1 \cite{howard2017mobilenets} for 200 epochs using a batch size of 128 with cutout and cosine learning rate decay. For ResNet-18, ResNet-50, we use $64 \times 64$ resolution images and transform the first convolution of the models to a $3 \times 3$ convolution. For MobileNetv1, we resize the original images to $224 \times 224$. 

We rename AUSAM as AUSAM-$\alpha$ according to the value of $\alpha$ when reporting the results.
Each experiment was repeated three times using different random seeds to calculate the standard deviation, and the results are reported as average. SAM\% represents the ratio of the training speed of a method to that of SAM. We implement AUSAM in Pytorch and train models on a single NVIDIA GeForce RTX 3090.

\begin{table}
\centering
\setlength{\tabcolsep}{1mm}{
\begin{tabular}{ccccc}
\toprule
\multirow{2}{*}{Methods} & \multicolumn{2}{c}{Resnet-18} & \multicolumn{2}{c}{Wideresnet-28-10} \\ \cline{2-5} 
                          & Accuracy       & SAM\%      & Accuracy          & SAM\%          \\ \midrule
SGD                       & 79.84\scriptsize{$\pm$0.45} & 171\% & 82.47\scriptsize{$\pm$0.30} & 195\%   \\
SAM                       & \bf{81.04\scriptsize{$\pm$0.29}} & 100\% & \underline{84.71\scriptsize{$\pm$0.21}} & 100\%   \\ \hline
AUSAM-0.4                       & 80.14\scriptsize{$\pm$0.29} & 155\% & 84.44\scriptsize{$\pm$0.31} & 202\%   \\
AUSAM-0.5                       & 80.67\scriptsize{$\pm$0.05} & 132\% & 84.62\scriptsize{$\pm$0.26} & 179\%   \\
AUSAM-0.6                       & 80.69\scriptsize{$\pm$0.15} & 119\% & \bf{84.73\scriptsize{$\pm$0.29}} & 152\%   \\
AUSAM-0.7                       & \underline{80.92\scriptsize{$\pm$0.17}} & 114\% & 84.66\scriptsize{$\pm$0.07} & 134\%   \\ \bottomrule
\end{tabular}}
\caption{Parameter Study of $\alpha$. The best accuracy is in bold and the second best is underlined.}
\label{table_alpha}
\end{table}

\begin{table*}
\center
\setlength{\tabcolsep}{4.3mm}{
\begin{tabular}{ccccc}
\toprule
& \multicolumn{2}{c}{\bf{CIFAR-10}} & \multicolumn{2}{c}{\bf{CIFAR-100}} \\ \midrule \midrule
\bf{ResNet-18}     & Accuracy $\uparrow$  & SAM\% $\uparrow$ & Accuracy $\uparrow$ & SAM\% $\uparrow$ \\ \midrule 
SGD                & 96.23\scriptsize{$\pm$0.11}                & 170\% 
                   & 79.84\scriptsize{$\pm$0.45}                & 171\% \\ 
SAM                & \bf{96.79\scriptsize{$\pm$0.03}}                & 100\%  
                   & \bf{81.04\scriptsize{$\pm$0.29}}    & 100\% \\ \midrule
LookSAM-5          & 96.47\scriptsize{$\pm$0.13}                & 115\%
                   & 80.48\scriptsize{$\pm$0.24}                & 113\% \\
ESAM               & 96.58\scriptsize{$\pm$0.13}                & 104\% 
                   & 80.47\scriptsize{$\pm$0.31}                & 105\% \\
ESAM\textsuperscript{1}
                   & 96.56\scriptsize{$\pm$0.08}                & 140\% 
                   & 80.41\scriptsize{$\pm$0.10}                & 140\% \\
SAF                & 96.15\scriptsize{$\pm$0.11}                & 147\% 
                   & 80.03\scriptsize{$\pm$0.43}                & 147\% \\
SAF\textsuperscript{2}
                   & 96.37\scriptsize{$\pm$0.02}                & 194\% 
                   & 80.06\scriptsize{$\pm$0.05}                & 192\% \\
MESA\textsuperscript{2}
                   & 96.24\scriptsize{$\pm$0.02}                & 168\% 
                   & 79.79\scriptsize{$\pm$0.09}                & 165\% \\ 
K-SAM\textsuperscript{3}
                   & 96.47\scriptsize{$\pm$0.05}                & 190\% 
                   & 79.00\scriptsize{$\pm$0.16}                & 189\% \\ 
G-RST\textsuperscript{4}
                   & 96.35\scriptsize{$\pm$0.10}                & / 
                   & 80.05\scriptsize{$\pm$0.18}                & / \\ 
                   \midrule
AUSAM-0.4          & 96.26\scriptsize{$\pm$0.10}                &  156\% 
                   & 80.14\scriptsize{$\pm$0.29}                & 155\% \\ 
AUSAM-0.5          & 96.52\scriptsize{$\pm$0.01}                &  132\% 
                   & 80.67\scriptsize{$\pm$0.05}                & 132\% \\   
AUSAM-0.6          & \underline{96.59\scriptsize{$\pm$0.14}}                & 123\% 
                   & \underline{80.69\scriptsize{$\pm$0.15}}    & 119\%  \\                      \midrule \midrule
\bf{WideResNet-28-10}    & Accuracy $\uparrow$ & SAM\% $\uparrow$  & Accuracy $\uparrow$ & SAM\% $\uparrow$ \\ \midrule
SGD                & 96.91\scriptsize{$\pm$0.06}                & 195\%  
                   & 82.47\scriptsize{$\pm$0.30}                & 195\% \\ 
SAM                & \underline{97.44\scriptsize{$\pm$0.04}}    & 100\%  
                   & \underline{84.71\scriptsize{$\pm$0.21}}                & 100\% \\ \midrule
LookSAM-5          & 97.13\scriptsize{$\pm$0.04}                & 154\%  
                   & 83.52\scriptsize{$\pm$0.09}                & 157\% \\
ESAM               & 97.41\scriptsize{$\pm$0.07}                & 144\% 
                   & \underline{84.71\scriptsize{$\pm$0.32}}    & 141\% \\
ESAM\textsuperscript{1}
                   & 97.29\scriptsize{$\pm$0.11}                & 139\% 
                   & 84.51\scriptsize{$\pm$0.01}                & 139\% \\
SAF                & 96.96\scriptsize{$\pm$0.09}                & 183\% 
                   & 82.57\scriptsize{$\pm$0.18}                & 182\% \\
SAF\textsuperscript{2} 
                   & 97.08\scriptsize{$\pm$0.15}                & 198\% 
                   & 83.81\scriptsize{$\pm$0.04}                & 197\% \\
MESA\textsuperscript{2}
                   & 97.16\scriptsize{$\pm$0.23}                & 168\% 
                   & 83.59\scriptsize{$\pm$0.24}                & 169\% \\
K-SAM\textsuperscript{3}
                   & 97.41\scriptsize{$\pm$0.05}                & 217\% 
                   & 84.05\scriptsize{$\pm$0.19}                & 204\% \\
G-RST\textsuperscript{4}
                   & 97.32\scriptsize{$\pm$0.05}                & / 
                   & 83.81\scriptsize{$\pm$0.15}                & / \\ 
                   \midrule
AUSAM-0.4          & 97.17\scriptsize{$\pm$0.03}                 & 204\%  
                   & 84.44\scriptsize{$\pm$0.31}                & 202\%  \\
AUSAM-0.5          & 97.3\scriptsize{$\pm$0.06}                 & 179\%  
                   & 84.62\scriptsize{$\pm$0.26}                & 179\%  \\      
AUSAM-0.6          & \bf{97.49\scriptsize{$\pm$0.05}}           & 152\% 
                   & \bf{84.73\scriptsize{$\pm$0.29}}           & 152\%  \\                   \midrule \midrule
\bf{PyramidNet-110}    & Accuracy $\uparrow$ & SAM\% $\uparrow$  & Accuracy $\uparrow$ & SAM\% $\uparrow$ \\ \midrule
SGD                & 97.14\scriptsize{$\pm$0.08}                & 195\%  
                   & 83.38\scriptsize{$\pm$0.21}                & 196\% \\
SAM                &97.69\scriptsize{$\pm$0.09}                 & 100\%  
                   &\bf{86.06\scriptsize{$\pm$0.16}}            & 100\% \\ \midrule
LookSAM-5          & 97.22\scriptsize{$\pm$0.05}                & 121\% 
                   & 83.76\scriptsize{$\pm$0.45}                & 126\% \\
ESAM               & 97.59\scriptsize{$\pm$0.17}                & 115\%  
                   & 85.32\scriptsize{$\pm$0.03}                & 116\% \\
ESAM\textsuperscript{1}
                   & \bf{97.81\scriptsize{$\pm$0.01}}           & 139\% 
                   & 85.56\scriptsize{$\pm$0.05}                & 138\% \\
SAF                & 96.96\scriptsize{$\pm$0.05}                & 178\%
                   & 83.66\scriptsize{$\pm$0.34}                & 181\% \\
SAF\textsuperscript{2}
                   & 97.34\scriptsize{$\pm$0.06}                & 202\% 
                   & 84.71\scriptsize{$\pm$0.01}                & 200\% \\
MESA\textsuperscript{2}
                   & 97.46\scriptsize{$\pm$0.09}                & 171\% 
                   & 84.73\scriptsize{$\pm$0.14}                & 171\% \\
K-SAM\textsuperscript{3}
                   & 97.62\scriptsize{$\pm$0.10}                & 195\% 
                   & 84.60\scriptsize{$\pm$0.22}                & 204\% \\
                   \midrule
AUSAM-0.4          & 97.51\scriptsize{$\pm$0.02}                &  203\% 
                   & 84.86\scriptsize{$\pm$0.25}                & 201\% \\ 
AUSAM-0.5          & 97.63\scriptsize{$\pm$0.04}                & 175\% 
                   & 85.68\scriptsize{$\pm$0.03}                & 176\%  \\ 
AUSAM-0.6          & \underline{97.72\scriptsize{$\pm$0.09}}                & 151\% 
                   & \underline{85.78\scriptsize{$\pm$0.06}}                & 153\%  \\ \bottomrule
\multicolumn{5}{l}{\small $1$ We report the results in \cite{du-2022-ESAM-ICLR}. }\\  % But failed to reproduce them using the officially released codes.              
\multicolumn{5}{l}{\small $2$ We report the results in \cite{du-2022-saf-NIPS}. } \\ % 
\multicolumn{5}{l}{\small $3$ We report the results in \cite{ni2022k}. } \\ %
\multicolumn{5}{l}{\small $4$ We report the results in \cite{zhao2022randomized}. } \\ %
\end{tabular}}
\caption{The results of the proposed method and the comparison methods on CIFAR-10 and CIFAR-100 dataset.
SAM\% indicate the ratio of corresponding method's training speed to SAM’s.
$\uparrow$ means that the larger the reported results are better.
The best accuracy is in bold and the second best is underlined.}
\label{table_result_cifar}
\end{table*}

\begin{table*}
\centering
\begin{tabular}{ccccccc}
\hline
& \multicolumn{2}{c}{ResNet-18}  & \multicolumn{2}{c}{ResNet-50} & \multicolumn{2}{c}{MobileNetv1}   \\ \hline
Methods    & Accuracy      & SAM\%     & Accuracy    & SAM\%  & Accuracy    & SAM\% \\ \hline
SGD       & 61.88\scriptsize{$\pm$0.31} & 198\% & 65.60\scriptsize{$\pm$0.51}  & 201\% & 58.12\scriptsize{$\pm$0.4}  & 189\% \\
SAM       & \underline{64.43\scriptsize{$\pm$0.08}} & 100\% & \bf{67.69\scriptsize{$\pm$0.10}}  & 100\%  & 58.49\scriptsize{$\pm$0.38} & 100\% \\ \hline
AUSAM-0.5 & 64.37\scriptsize{$\pm$0.03} & 191\% & 66.60\scriptsize{$\pm$0.21}  & 195\% & \underline{58.55\scriptsize{$\pm$0.23}} & 185\% \\
AUSAM-0.6 &  \bf{64.49\scriptsize{$\pm$0.39}} & 151\% & \underline{66.65\scriptsize{$\pm$0.34}} & 166\% & \bf{58.94\scriptsize{$\pm$0.24}} & 161\% \\ \hline
\end{tabular}
\caption{The results of SGD, SAM and AUSAM on Tiny-ImageNet. The best accuracy is in bold and the second best is underlined.}
\label{table_result_tiny}
\end{table*}

\subsection{Parameter Studies}
We study the effect of $\alpha$ on accuracy and optimization efficiency using ResNet-18 and WideResNet-28-10 on the CIFAR-100 dataset.  The corresponding results are summarized in Tab. \ref{table_alpha}. 

$\alpha$ determines the number of samples in each mini-batch, significantly influencing both the efficiency and accuracy of model training. As $\alpha$ decreases, the acceleration ratio of AUSAM increases gradually. This is because, as the number of samples selected in each mini-batch decreases, forward and backward propagation becomes more and more efficient. In theory, when $\alpha=0.5$, the speed of AUSAM can nearly match that of the base optimizer, such as SGD. However, the time spent on other operations, such as data loading, is non-negligible compared to the time spent on forward and backward, AUSAM cannot accelerate these operations. Therefore, it cannot achieve the same speed as SGD. For ResNet-18, due to the smaller model size, the optimization time constitutes a smaller proportion of the overall training time, leading to a relatively low overall acceleration ratio. For WideResNet, the network requires more time for forward and backward, which allows AUSAM to fully utilize its performance. When $\alpha=0.4$, the training speed matches that of SGD, and the performance is significantly improved compared to SGD.

\subsection{Comparison to SOTAs}
The experimental results on CIFAR-10 and CIFAR-100 are presented in Tab.~\ref{table_result_cifar}.
We observe that AUSAM achieves significantly higher accuracy compared to SGD. 
Furthermore, in comparison to LookSAM and ESAM, our results outperform theirs in most cases. 
This demonstrates that AUSAM can successfully preserve the model's generalization ability during the training process.
This is probably because AUSAM essentially reduces the batch size, which helps to improve the generalization of the model \cite{he－2019－control}. 
Although the total number of iterations for the entire training process did not increase, the selected samples accelerated the model's training.
AUSAM also achieves superior results compared to SAF and MESA.
This is because SAF and MESA optimize sharpness based on multiple previous iterations up to the current one, which differs from the sharpness optimization at the current iteration alone.

When $\alpha=0.5$, these experiments demonstrate that AUSAM exhibits about 70\% faster training than SAM while achieving comparable accuracy.
For ResNet-18, the benefits of AUSAM are not fully apparent due to the inherently shorter duration required for both forward and backward during the optimization process. 
With larger models like WideResNet-28-10 and PyramidNet-110, AUSAM's benefits become more pronounced, resulting in faster training.
When $\alpha=0.4$, for WideResNet-28-10 and PyramidNet-110, AUSAM's speed is close to that of the base optimizer (SGD), and its performance is superior to the base optimizer (SGD). Compared to SAF and K-SAM, AUSAM performs even better.

\begin{table*}
\centering
\begin{tabular}{ccccccccccc}
\hline
           & Hea   & Sho & Elb  & Wri  & Hip    & Kne   & Ank  & Mean  & SAM\% \\ \hline
Adam       & 96.8 &  95.8  & 89.7 &    84.3    &  88.7 & 85.1  & 81.4 & 89.3 &  183\%                  \\
SAM  &  97.0  &  95.7  & 89.6  & 84.2   & 88.4  &  85.5  &  81.9 & \underline{89.4}  &  100\%  \\
AUSAM & 96.9 & 95.7   & 89.4 & 84.8 & 89.1 & 85.7 & 81.3 & \bf{89.5} & 144\%         \\ \hline
\end{tabular}
\caption{Results of training SimCC on MP$\mathrm{\uppercase\expandafter{\romannumeral2}}$ with Adam,SAM and AUSAM.}
\label{table_hpe}
\end{table*}

The experimental results on Tiny-ImageNet are presented in Tab.~\ref{table_result_tiny}. 
Due to the larger image resolution of Tiny-ImageNet compared to CIFAR-10 and CIFAR-100, the network requires more time for both forward and backward computations, leading to a higher acceleration ratio for AUSAM. With $\alpha=0.5$, AUSAM achieves near SGD speed while sustaining superior accuracy in comparison to SGD for these three models. On ResNet-18 and MobileNetv1, AUSAM achieves results comparable to SAM. 

\subsection{Comparison to Dynamic Data Pruning}
Data pruning aims to achieve lossless performance while minimizing overall costs by pruning less informative data at each epoch and training with the remaining data. We apply the advanced dynamic data pruning method InfoBatch \cite{qin-2024-infobatch} to the CIFAR-100 and conduct experiments using SGD and SAM. SAM+Info* represents the use of InfoBatch during the first loss calculation in SAM, while SAM+Info indicates its use during the second loss calculation in SAM. The results are presented in Tab.~\ref{table_infobatch}. 
Although InfoBatch improves optimization efficiency, it does not guarantee lossless optimization performance for SAM on CIFAR-100. Most dynamic data pruning methods struggle to effectively leverage the two losses before and after perturbation in SAM. Typically, these methods select samples with larger training losses directly, but such samples may not necessarily have larger losses in the parameter neighborhood space, which contradicts SAM's optimization principle.

\begin{table}
\centering
%\vspace{-10pt}
\begin{tabular}{lcc}
\toprule
Methods   & Accuracy   & SAM\%   \\ \midrule
SGD       & 79.79\scriptsize{$\pm$0.14} & 177\%  \\
SAM       & \bf{81.24}\scriptsize{$\pm$0.10} & 100\%  \\ \midrule
SGD+Info & 79.21\scriptsize{$\pm$0.24} & 197\% \\
SAM+Info* & 79.68\scriptsize{$\pm$0.26} & 137\% \\
SAM+Info & 79.57\scriptsize{$\pm$0.09} & 140\% \\ \midrule
SAM+Random & \underline{80.56\scriptsize{$\pm$0.19}} & 120\% \\ \bottomrule
\end{tabular}
\caption{The Results of using InfoBatch to accelerate SAM.}
\label{table_infobatch}
%\vspace{-10pt}
\end{table}

\begin{table}[h]
\center
\begin{tabular}{ccccc}
\toprule
           & \multicolumn{2}{c}{ResNet-18}  & \multicolumn{2}{c}{MobileNets} \\ \midrule
Methods    & Accuracy      & SAM\%     & Accuracy    & SAM\%   \\ \hline
Full prec. & 88.72     & \textbackslash{}  & 85.81      & \textbackslash{}           \\
SGD    & 88.86\scriptsize{$\pm$0.18}     & 174\%    & 84.04\scriptsize{$\pm$0.13}   & 153\%          \\
SAM    & \bf{89.75}\scriptsize{$\pm$0.21}     & 100\%     & \underline{84.72\scriptsize{$\pm$0.11}}   & 100\%           \\
AUSAM   & \underline{89.63\scriptsize{$\pm$0.10}}     & 130\%    & \bf{84.91\scriptsize{$\pm$0.19}}   & 117\%          \\ \bottomrule
\end{tabular}%}
\caption{Results of QAT with SGD, SAM and AUSAM on the Cifar-10.}
\label{table_QAT}
\end{table}

\subsection{Application to Human Pose Estimation}
2D Human Pose Estimation (HPE) aims to localize body joints from a single image. We apply SAM and AUSAM to HPE to evaluate their general applicability, using the SimCC method \cite{li2022simcc} for validation. The key idea of SimCC is to treat human pose estimation as two classification tasks, one for vertical and one for horizontal coordinates. We use SGD, SAM, and AUSAM as optimizers for the SimCC, and conducted experiments on the MP$\mathrm{\uppercase\expandafter{\romannumeral2}}$ dataset \cite{andriluka14cvpr}. We follow the evaluation procedure in  \cite{li2022simcc}, and conduct experiments with an image resolution of $256 \times 256$. In the experiments, both SAM and AUSAM use the Adam optimizer as their base optimizer, with a batch size set to 64. The results are shown in Tab.~\ref{table_hpe}. It can be found that both SAM and AUSAM obtain better performance than Adam. In terms of speed, AUSAM is faster than SAM but slower than Adam. This suggests that AUSAM is applicable across a broad spectrum of scenarios.

\subsection{Application to Quantization-Aware Training}
Neural network quantization reduces computational demands by lowering weight and activation precision, enabling efficient deployment on edge devices without compromising model performance~\cite{jacob-2018-quantization,esser2020learned,wei2021qdrop,nagel2022overcoming}. We employ AUSAM as the optimizer for QAT \cite{jacob-2018-quantization} to demonstrate its broader applicability. We applied  the SGD, SAM, and AUSAM algorithms to quantize the parameters of the ResNet-18 and MobileNetV1 models to W4A4 on the CIFAR-10 dataset. The results are presented in Tab.~\ref{table_QAT}. Although the time required for both forward and backward is significantly reduced after model quantization, experimental results show that AUSAM can still accelerate the optimization speed while maintaining performance.

\section{Conclusions}
The findings of this study reveal that during the optimization process of SAM, training with samples that have larger gradient norms from each mini-batch effectively preserves the model's generalization capability without significant deterioration. Based on this observation, this paper proposes an asymptotic unbiased sampling strategy aimed at accelerating SAM. Integrating this strategy into SAM enhances optimization speed by approximately 70\% while maintaining the model's generalization capability. AUSAM samples a subset of data from the mini-batch based on the estimated average gradient norm before each optimization iteration, significantly reduces the forward and backward time. This is particularly beneficial for large models that require a significant amount of time for forward and backward. Experiments on the CIFAR-10, CIFAR-100, and Tiny-ImageNet datasets show that AUSAM not only matches SAM in model generalization but also significantly surpasses it in optimization speed. Additionally, we applied AUSAM to human pose estimation and model quantization tasks. The results show that AUSAM enhances optimization speed while preserving performance, demonstrating its broad applicability.

\section*{Acknowledgements}
The work was supported by the National Key Research and Development Program of China (Grant No. 2023YFC3306401). This research was also supported by Zhejiang Provincial Natural Science Foundation of China under Grant No. LD24F020007, Beijing Natural Science Foundation L223024 and L244043, National Natural Science Foundation of China under Grant 61872333, 62076016, “One Thousand Plan” projects in Jiangxi Province Jxsq2023102268.

\bibliography{aaai25}

\end{document}

\maketitle

\appendix

\section{Main proof}
\label{sec:main_proof}

\subsection{Proof of Theorem 1}
\begin{proof}
Suppose we select $N$ samples from a mini-batch $\mathbf{B}$ of size $K$, denoting them as $\hat{\mathbf{B}}$ ($\hat{{x}_i} \in \hat{\mathbf{B}}$, $i=1,2,...,N$), and represent the remaining set of $M(M=K-N)$ samples as $\check{\mathbf{B}}$ ($\check{{x}_i} \in \check{\mathbf{B}}$, $i=1,2,...,M$). Let $\bm{\varepsilon}_\mathbf{B}$ and $\bm{\varepsilon}_{\hat{\mathbf{B}}}$ represent the parameter perturbations obtained through the set $\mathbf{B}$ and $\hat{\mathbf{B}}$, respectively.
 The difference in the Perturbation loss and Training loss between the calculations using a Full mini-batch (PTF) and those using Sampled samples (PTS) is as follows:
\begin{equation}
\begin{aligned}
&\left|\left[ {{L_\mathbf{B}}(\mathbf{w} + {{\hat {\bm{\varepsilon}} }_\mathbf{B}}) - {L_\mathbf{B}}(\mathbf{w})} \right] - \left[ {{L_{\hat{\mathbf{B}}}}(\mathbf{w} + {{\hat {\bm{\varepsilon}} }_{\hat {\mathbf{B}}}}) - {L_{\hat {\mathbf{B}}}}(\mathbf{w})} \right]\right|\\
 &= \left|{{\hat {\bm{\varepsilon}} }_\mathbf{B}}^{\rm T}{\nabla _w}{L_\mathbf{B}}(\mathbf{w}) + O({{\hat {\bm{\varepsilon}} }_\mathbf{B}}^2) - {{\hat {\bm{\varepsilon}} }_{\hat {\mathbf{B}}}}^{\rm T}{\nabla _w}{L_{\hat {\mathbf{B}}}}(\mathbf{w}) - O({{\hat {\bm{\varepsilon}} }_{\hat {\mathbf{B}}}}^2)\right|\\
 &= \big|\rho \frac{{{\nabla _\mathbf{w}}{L_\mathbf{B}}{{(\mathbf{w})}^{\rm T}}}}{{\left\| {{\nabla _\mathbf{w}}{L_\mathbf{B}}(\mathbf{w})} \right\|}}{\nabla _\mathbf{w}}{L_\mathbf{B}}(\mathbf{w}) - \rho \frac{{{\nabla _\mathbf{w}}{L_{\hat {\mathbf{B}}}}{{(\mathbf{w})}^{\rm T}}}}{{\left\| {{\nabla _\mathbf{w}}{L_{\hat {\mathbf{B}}}}(\mathbf{w})} \right\|}}{\nabla _\mathbf{w}}{L_{\hat {\mathbf{B}}}}(\mathbf{w})\\
 &+ (O({{\hat {\bm{\varepsilon}} }_\mathbf{B}}^2) - O({{\hat {\bm{\varepsilon}} }_{\hat {\mathbf{B}}}}^2))\big|\\
 &= \left|\rho \left\| {{\nabla _\mathbf{w}}{L_\mathbf{B}}(\mathbf{w})} \right\| - \rho \left\| {{\nabla _\mathbf{w}}{L_{\hat {\mathbf{B}}}}(\mathbf{w})} \right\| + (O({{\hat {\bm{\varepsilon}} }_\mathbf{B}}^2) - O({{\hat {\bm{\varepsilon}} }_{\hat {\mathbf{B}}}}^2))\right|\\
 &= \big|\rho \big\| {\frac{1}{N}\sum\limits_{i = 1}^N {{\nabla _\mathbf{w}}{L_{{\hat{{x}_i}}}}(\mathbf{w})}  + \frac{1}{M}\sum\limits_{i = 1}^M {{\nabla _\mathbf{w}}{L_{{\check{{x}_i}}}}(\mathbf{w})} } \big\| \\
 &- \rho \big\| {\frac{1}{N}\sum\limits_{i = 1}^N {{\nabla _\mathbf{w}}{L_{{\hat{{x}_i}}}}(\mathbf{w})} } \big\| + (O({{\hat {\bm{\varepsilon}} }_\mathbf{B}}^2) - O({{\hat {\bm{\varepsilon}} }_{\hat {\mathbf{B}}}}^2))\big|\\
 &\le \rho \big\| {\frac{1}{M}\sum\limits_{i = 1}^M {{\nabla _\mathbf{w}}{L_{{\check{{x}_i}}}}(\mathbf{w})} } \big\| + \left|(O({{\hat {\bm{\varepsilon}} }_\mathbf{B}}^2) - O({{\hat {\bm{\varepsilon}} }_{\hat {\mathbf{B}}}}^2))\right|\\
 &\le \frac{\rho }{M}\sum\limits_{i = 1}^M {\left\| {{\nabla _\mathbf{w}}{L_{{\check{{x}_i}}}}(\mathbf{w})} \right\|}  + \left|(O({{\hat {\bm{\varepsilon}} }_\mathbf{B}}^2) - O({{\hat {\bm{\varepsilon}} }_{\hat {\mathbf{B}}}}^2))\right|
\end{aligned}
\end{equation}

\end{proof}

\subsection{Proof of Theorem 2}
\begin{assumption}\label{assumpion:smooth}
(Smoothness). $L(\mathbf{w})$ is $\tau$-Lipschitz smooth in $\mathbf{w}$, i.e., $\left\| {\nabla L(\mathbf{w}) - \nabla L(\mathbf{v})} \right\| \le \tau \left\| {\mathbf{w} - \mathbf{v}} \right\|$.
\end{assumption}
\begin{assumption}\label{assumpion:Bounded_g}
(Bounded gradients). By the assumption that an upper bound is exists on the gradient of sampled set $\hat{\mathbf{B}}$ from each mini-batch. There exists $G > 0$ for $\hat{\mathbf{B}}$ such that $\mathbb{E}\left[ {\left\| {\nabla L_{\hat{\mathbf{B}}}(\mathbf{w})} \right\|} \right] \le G$.
\end{assumption}
\begin{assumption} \label{assumpion:bound_var}
(Bounded variance of stochastic gradients). Given the training set $\mathbf{D}$ and a sampled set $\hat{\mathbf{B}} \in \mathbf{D}$. There exists $\sigma \ge 0$, the variance of the sampled set of size $\alpha K$ is bounded by $\mathbb{E}\left[ {{{\left\| {\nabla {L_{\hat{\mathbf{B}}}}(\mathbf{w}) - \nabla {L_\mathbf{D}}(\mathbf{w})} \right\|}^2}} \right] \le \frac{\sigma ^2}{\alpha K}$.
\end{assumption}
\begin{proof}
Suppose Assumption \ref{assumpion:smooth} and \ref{assumpion:Bounded_g} hold. The deviation between the average gradient norm of sample $\mathbf{x}_i$ over $T-1$ epochs and the gradient norm at the $T$-th epoch is defined as follows:
\begin{align}
&\left| {\left\| {\nabla {L_{{B_i}}}({\mathbf{w}_T})} \right\| - \frac{1}{{T - 1}}\sum\limits_{t = 1}^{T - 1} {\left\| {{\nabla _\mathbf{w}}{L_{{B_i}}}({\mathbf{w}_t})} \right\|} } \right|\\ \nonumber
&= \frac{1}{{T - 1}}\left| {\sum\limits_{t = 1}^{T - 1} {\left( {\left\| {\nabla {L_{{B_i}}}({\mathbf{w}_T})} \right\| - \left\| {{\nabla _\mathbf{w}}{L_{{B_i}}}({\mathbf{w}_t})} \right\|} \right)} } \right|\\ \nonumber
&\le \frac{1}{{T - 1}}\sum\limits_{t = 1}^{T - 1} {\left| {\left\| {\nabla {L_{{B_i}}}({\mathbf{w}_T})} \right\| - \left\| {{\nabla _\mathbf{w}}{L_{{B_i}}}({\mathbf{w}_t})} \right\|} \right|} \\ \nonumber
&\le \frac{1}{{T - 1}}\sum\limits_{t = 1}^{T - 1} {\left\| {\nabla {L_{{B_i}}}({\mathbf{w}_T}) - {\nabla _\mathbf{w}}{L_{{B_i}}}({\mathbf{w}_t})} \right\|} \\ \nonumber
&\le \frac{\tau }{{T - 1}}\sum\limits_{t = 1}^{T - 1} {\left\| {{\mathbf{w}_T} - {\mathbf{w}_t}} \right\|} \\ \nonumber
&= \frac{\tau }{{T - 1}}\left( {\left\| {{\mathbf{w}_T} - {\mathbf{w}_1}} \right\| + \left\| {{\mathbf{w}_T} - {\mathbf{w}_2}} \right\| +  \cdot  \cdot  \cdot  + \left\| {{\mathbf{w}_T} - {\mathbf{w}_{T - 1}}} \right\|} \right)\\ \nonumber
&\le \frac{\tau }{{T - 1}}\sum\limits_{t = 1}^{T - 1} {\sum\limits_{j = t + 1}^T {\left\| {{\mathbf{w}_j} - {\mathbf{w}_{j - 1}}} \right\|} } .
\end{align}
Suppose that the parameters are updated $N$ iterations between and $\mathbf{w}_j$ and $\mathbf{w}_{j - 1}$. Based on Assumption~\ref{assumpion:Bounded_g}, it follows that:
\begin{equation}
\begin{aligned}
&\frac{\tau }{{T - 1}}\sum\limits_{t = 1}^{T - 1} {\sum\limits_{j = t + 1}^T {\left\| {{\mathbf{w}_j} - {\mathbf{w}_{j - 1}}} \right\|} } \\
&= \frac{\tau }{{T - 1}}\sum\limits_{t = 1}^{T - 1} {(T-t)\left\| {N{\eta _t}\nabla {L_\mathbf{B}}(\mathbf{w})} \right\|}\\
&= \frac{{\tau NG}}{{T - 1}}\sum\limits_{t = 1}^{T - 1} {{\eta _t}(T-t)} .
\end{aligned}
\end{equation}
Let ${\eta _t} = \frac{{{\eta _0}}}{{ t^2 }}$, we have:
\begin{equation}
\begin{aligned}
&\frac{{\tau NG}}{{T - 1}}\sum\limits_{t = 1}^{T - 1} {{\eta _t}(T-t)}
= \frac{{\tau NG}}{{T - 1}}\sum\limits_{t = 1}^{T - 1} {\frac{{{\eta _0}}}{{t^2 }}(T-t)}\\
&= \frac{{\tau NG{\eta _0}}}{{T - 1}}\sum\limits_{t = 1}^{T - 1} {\frac{T-t}{{t^2 }}}
\le \frac{{\tau NG{\eta _0}}}{{T - 1}}\sum\limits_{t = 1}^{T - 1} {\frac{T-1}{{t^2 }}}
\le \frac{{\tau {\eta _0} \pi^2 NG}}{6}.
\end{aligned}
\end{equation}

\end{proof}

\subsection{Proof of Theorem 3}
%\begin{assumption}
%(Bounded variance of stochastic gradients). There exists $\sigma \ge 0$, the variance of a mini-batch of size $K$ is bounded by $\mathbb{E}\left[ {{{\left\| {\nabla {L_\mathbf{B}}(\mathbf{w}) - \nabla {L_\mathbf{D}}(\mathbf{w})} \right\|}^2}} \right] \le \frac{\sigma ^2}{K}$.
%\end{assumption}
\begin{proof}
Assume that the true gradient of all data is ${{\mathbf{g}}_t} = \nabla L_{\mathbf{D}}({\mathbf{w}_t})$. Let $K$ be the mini-batch size. 
In the stage where sam computes the perturbation, the gradient of sampling set $\mathbf{B}_t$ is ${\hat{\mathbf{g}}_t} = \frac{1}{K}\sum\nolimits_{{\mathbf{x}_i} \in {\mathbf{B}_t}} {\nabla L_{{\mathbf{x}_i}}({\mathbf{w}_t})}$. 
Let us define the weights after perturbation as $\mathbf{w}_t^{adv} = {\mathbf{w}_t} + \rho {\hat{\mathbf{g}}_t}$.
In the gradient descent phase, the gradient of sampling set is ${\hat{\mathbf{h}}_t} = \frac{1}{K}\sum\nolimits_{{\mathbf{x}_i} \in {\mathbf{B}_t}} {\nabla L_{\mathbf{x}_i}({\mathbf{w}_t} + \rho {\hat{\mathbf{g}}_t})}$.
The weight update process in AUSAM can be defined as follows:
\begin{equation}
\begin{aligned}
{\mathbf{w}_{t + 1}} = {\mathbf{w}_t} - \eta_t {\hat{\mathbf{h}}_t},
\end{aligned}
\end{equation}
where $\eta_t$ is the learning rate. Using the smoothness of the function $L$ (Assumption \ref{assumpion:smooth}), we obtain:
\begin{equation}
\begin{aligned}
&L({\mathbf{w}_{t + 1}}) \le L({\mathbf{w}_t}) + {\mathbf{g}_t}^ \top ({\mathbf{w}_{t + 1}} - {\mathbf{w}_t}) + \frac{\tau }{2}{\left\| {{\mathbf{w}_{t + 1}} - {\mathbf{w}_t}} \right\|^2}\\ \nonumber
&= L({\mathbf{w}_t}) - \eta {\mathbf{g}_t}^ \top {{\hat{\mathbf{h}}}_t} + \frac{{\tau {\eta ^2}}}{2}{\left\| {{{\hat{\mathbf{h}}}_t}} \right\|^2} \\ \nonumber
&= L({\mathbf{w}_t}) - \eta {\mathbf{g}_t}^ \top {{\hat{\mathbf{h}}}_t} + \frac{{\tau {\eta ^2}}}{2}\left( {{{\left\| {{{{\rm{\hat{\mathbf{h}}}}}_t} - {\mathbf{g}_t}} \right\|}^2} - {{\left\| {{\mathbf{g}_t}} \right\|}^2} + 2{\mathbf{g}_t}^ \top {{\hat{\mathbf{h}}}_t}} \right)\\ \nonumber
&= L({\mathbf{w}_t}) + \frac{{\tau {\eta ^2}}}{2}{\left\| {{{{\rm{\hat{\mathbf{h}}}}}_t} - {\mathbf{g}_t}} \right\|^2} - \frac{{\tau {\eta ^2}}}{2}{\left\| {{\mathbf{g}_t}} \right\|^2} - \eta \left( {1 - \tau \eta } \right){\mathbf{g}_t}^ \top {{{\rm{\hat{\mathbf{h}}}}}_t}\\ \nonumber
&\le L({\mathbf{w}_t}) - \frac{{\tau {\eta ^2}}}{2}{\left\| {{\mathbf{g}_t}} \right\|^2} + \frac{{\tau {\eta ^2}}}{2}\left( {2{{\left\| {{{{\rm{\hat{\mathbf{h}}}}}_t} - {{\hat{\mathbf{g}}}_t}} \right\|}^2} + 2{{\left\| {{{\hat{\mathbf{g}}}_t} - {\mathbf{g}_t}} \right\|}^2}} \right)\\
&- \eta \left( {1 - \tau \eta } \right){\mathbf{g}_t}^ \top {{{\rm{\hat{\mathbf{h}}}}}_t}\\ \nonumber
%&\le L({\mathbf{w}_t}) - \frac{{\tau {\eta ^2}}}{2}{\left\| {{\mathbf{g}_t}} \right\|^2} + \tau {\eta ^2}{\left\| {{{{\rm{\hat{\mathbf{h}}}}}_t} - {{\hat{\mathbf{g}}}_t}} \right\|^2} + \tau {\eta ^2}{\left\| {{{\hat{\mathbf{g}}}_t} - {\mathbf{g}_t}} \right\|^2}\\
%&- \eta \left( {1 - \tau \eta } \right){\mathbf{g}_t}^ \top {{{\rm{\hat{\mathbf{h}}}}}_t}\\
&\le L({\mathbf{w}_t}) - \frac{{\tau {\eta ^2}}}{2}{\left\| {{\mathbf{g}_t}} \right\|^2} + {\tau ^3}{\eta ^2}{\rho ^2}{\left\| {{{\hat{\mathbf{g}}}_t}} \right\|^2} + \tau {\eta ^2}{\left\| {{{\hat{\mathbf{g}}}_t} - {\mathbf{g}_t}} \right\|^2}\\
&- \eta \left( {1 - \tau \eta } \right){\mathbf{g}_t}^ \top {{{\rm{\hat{\mathbf{h}}}}}_t}, \label{equ:T3_1}
\end{aligned}
\end{equation}
where \eqref{equ:T3_1} follows from ${\left\| {{{{\rm{\hat{\mathbf{h}}}}}_t} - {{\hat{\mathbf{g}}}_t}} \right\|^2} \le {\tau ^2}{\left\| {{\mathbf{w}_t} + \rho {{\hat{\mathbf{g}}}_t} - {\mathbf{w}_t}} \right\|^2} = {\tau ^2}{\rho ^2}{\left\| {{{\hat{\mathbf{g}}}_t}} \right\|^2}$. 

\begin{lemma}\label{lem:andri}
(Andriushchenko \& Flammarion (2022)). Under Assumptions \ref{assumpion:smooth} and \ref{assumpion:bound_var} for all $t$ and $\rho>0$, we have:
\begin{equation}
\begin{aligned}
\mathbb E[{\mathbf{g}_t}^{\top}{\hat{\mathbf{h}}_t}] \ge (\frac{1}{2} - \rho \tau ){\left\| {{\mathbf{g}_t}} \right\|^2} - \frac{{{\rho ^2}{\tau ^2}{\sigma ^2}}}{{2\alpha K}}.
\end{aligned}
\end{equation}
\end{lemma}
By Lemma \ref{lem:andri}, taking the expectation, we obtain:
\begin{equation}
\begin{aligned}
&\mathbb{E}L({\mathbf{w}_{t + 1}}) \le \mathbb{E}L({\mathbf{w}_t}) - \frac{{\tau {\eta ^2}}}{2}\mathbb{E}{\left\| {{\mathbf{g}_t}} \right\|^2} + {\tau ^3}{\eta ^2}{\rho ^2}\mathbb{E}{\left\| {{{\hat{\mathbf{g}}}_t}} \right\|^2} \\
& + \tau {\eta ^2}\mathbb{E}{\left\| {{{\hat{\mathbf{g}}}_t} - {\mathbf{g}_t}} \right\|^2} - \eta \left( {1 - \tau \eta } \right)\left( {(\frac{1}{2} - \rho \tau )\mathbb{E}{{\left\| {{\mathbf{g}_t}} \right\|}^{^2}} - \frac{{{\rho ^2}{\tau ^2}{\sigma ^2}}}{{2\alpha K}}} \right)\\
&= \mathbb{E}L({\mathbf{w}_t}) - \frac{{\tau {\eta ^2}}}{2}\mathbb{E}{\left\| {{\mathbf{g}_t}} \right\|^2} + {\tau ^3}{\eta ^2}{\rho ^2}\left( {\frac{{{\sigma ^2}}}{\alpha K} + \mathbb{E}{{\left\| {{\mathbf{g}_t}} \right\|}^2}} \right)\\ & + \frac{{{\sigma ^2}}}{\alpha K}\tau {\eta ^2} - \eta \left( {1 - \tau \eta } \right)(\frac{1}{2} - \rho \tau )\mathbb{E}{\left\| {{\mathbf{g}_t}} \right\|^{^2}} + \eta \left( {1 - \tau \eta } \right)\frac{{{\rho ^2}{\tau ^2}{\sigma ^2}}}{{2\alpha K}}\\
&= \mathbb{E}L({\mathbf{w}_t}) - \frac{{\tau {\eta ^2}}}{2}(1 - 2{\tau ^2}{\rho ^2})\mathbb{E}{\left\| {{\mathbf{g}_t}} \right\|^2} + \eta \left( {1 - \tau \eta } \right)\frac{{{\rho ^2}{\tau ^2}{\sigma ^2}}}{{2\alpha K}} \\ & + \tau {\eta ^2}\left( {1 + {\tau ^2}{\rho ^2}} \right)\frac{{{\sigma ^2}}}{\alpha K} - \eta \left( {1 - \tau \eta } \right)(\frac{1}{2} - \rho \tau )\mathbb{E}{\left\| {{\mathbf{g}_t}} \right\|^{^2}}\\
&= \mathbb{E}L({\mathbf{w}_t}) - \frac{{\tau {\eta ^2}}}{2}(1 - 2{\tau ^2}{\rho ^2})\mathbb{E}{\left\| {{\mathbf{g}_t}} \right\|^2} + \eta \left( {1 - \tau \eta } \right)\frac{{{\rho ^2}{\tau ^2}{\sigma ^2}}}{{2\alpha K}} \\ & + {\eta ^2}\left( {1 + {\tau ^2}{\rho ^2}} \right)\frac{{\tau {\sigma ^2}}}{\alpha K} - \eta \left( {1 - \tau \eta } \right)(\frac{1}{2} - \rho \tau )\mathbb{E}{\left\| {{\mathbf{g}_t}} \right\|^{^2}}.
\end{aligned}
\end{equation}
For $\eta  \le \frac{1}{\tau }$ and $\rho  \le \frac{1}{{4\tau }}$, summing over $T$ on both sides, we have:
\begin{equation}
\begin{aligned}
&\frac{1}{T}\mathbb{E}\left[ \sum\limits_{t = 1}^T {{{\left\| {{\mathbf{g}_t}} \right\|}^2}} \right]  \le \frac{{L[({\mathbf{w}_0})] - \mathbb{E}[L({\mathbf{w}_T})]}}{{\eta \sum\limits_{t = 1}^T {\left( {\frac{{\tau \eta }}{2}(1 - 2{\rho ^2}{\tau ^2}) + (1 - \tau \eta )(\frac{1}{2} - \rho \tau )} \right)} }} \\ 
& + \frac{{\sum\limits_{t = 1}^T ({2{\eta ^2}\left( {1 + {\tau ^2}{\rho ^2}} \right) + \eta \left( {1 - \tau \eta } \right){\rho ^2}\tau }) }}{{\eta \sum\limits_{t = 1}^T {\left( {\frac{{\tau \eta }}{2}(1 - 2{\rho ^2}{\tau ^2}) + (1 - \tau \eta )(\frac{1}{2} - \rho \tau )} \right)} }}\frac{{\tau {\sigma ^2}}}{{2\alpha K}}\\
& \le \frac{{16}}{{7\eta T}}\left( {L[({\mathbf{w}_0})] - E[L({\mathbf{w}_T})]} \right) + \frac{{17\eta \tau {\sigma ^2}}}{{7\alpha K}}.
\end{aligned}
\end{equation}

\end{proof}

\subsection{Proof of Theorem 4}
\begin{proof}
Assuming all samples $\mathbf{x}$ from $\mathbf{D}$ are drawn from continuous distribution $q(\mathbf{x})$. $p^t_\mathbf{x}$ is the probability that sample $\mathbf{x}$ is sampled in the $t$-th epoch. The perturbation of SAM is assumed to be the theoretically true perturbation, that is, the perturbation calculated with the whole data set $D$, ${\bm{\varepsilon}_\mathbf{D}} = \rho \frac{{{\nabla _\mathbf{w}}{L_D}({\mathbf{w}_t})}}{{\left\| {{\nabla _\mathbf{w}}{L_D}({\mathbf{w}_t})} \right\|}}$. The difference between the expected loss of full dataset and the expected loss of sampled subsets $\hat{\mathbf{D}}$ can be expressed as follows:
% \begin{equation}\label{equ:P1_1}
% \begin{aligned}
% &\mathop {\mathbb{E}}\limits_{\mathbf{x} \in D} \left[ {\mathop {\max }\limits_{\left\| {{\bm{\varepsilon} _\mathbf{x}}} \right\| \le \rho } [{L_\mathbf{x}}({\mathbf{w}_t} + {\bm{\varepsilon}_\mathbf{x}}) - {L_\mathbf{x}}({\mathbf{w}_t})]} \right] - \mathop {\mathbb{E}}\limits_{\mathbf{x} \in \hat D} \left[ {\mathop {\max }\limits_{\left\| {{\bm{\varepsilon}_\mathbf{x}}} \right\| \le \rho } [{L_\mathbf{x}}({\mathbf{w}_t} + {\bm{\varepsilon}_\mathbf{x}}) - {L_\mathbf{x}}({\mathbf{w}_t})]} \right]\\
% &= \rho \mathop {\rm{E}}\limits_{\mathbf{x} \in D} \left[ {\left\| {{\nabla _\mathbf{w}}{L_\mathbf{x}}(\mathbf{w})} \right\|} \right] - \rho \mathop {\rm{E}}\limits_{{\mathbf{x}} \in \hat D} \left[ {\left\| {{\nabla _\mathbf{w}}{L_\mathbf{x}}(\mathbf{w})} \right\|} \right]\\
% &= \rho \int_{\mathbf{x}} {\left\| {{\nabla _\mathbf{w}}{L_{\mathbf{x}}}(\mathbf{w})} \right\|q({\mathbf{x}})} d{\mathbf{x}} - \rho \int_{\mathbf{x}} {{p_\mathbf{x}^t}\left\| {{\nabla _\mathbf{w}}{L_{\mathbf{x}}}(\mathbf{w})} \right\|q({\mathbf{x}})} d{\mathbf{x}}\\
% &= \rho \int_{\mathbf{x}} {(1 - {p_\mathbf{x}^t})\left\| {{\nabla _\mathbf{w}}{L_{\mathbf{x}}}(\mathbf{w})} \right\|q({\mathbf{x}})} d{\mathbf{x}}
% \end{aligned}
% \end{equation}
\begin{small}
\begin{equation}
\begin{aligned}
&\bigg| \mathop {\rm{E}}\limits_{\mathbf{x} \in \mathbf{D}} \big[ {\mathop {\max }\limits_{\left\| {{\bm{\varepsilon} _\mathbf{D}}} \right\| \le \rho } [{L_\mathbf{x}}({\mathbf{w}_t} + {\bm{\varepsilon} _\mathbf{D}}) - {L_\mathbf{x}}({\mathbf{w}_t})]} \big]\\
&\quad\quad- \mathop {\rm{E}}\limits_{{\mathbf{x}} \in \hat{\mathbf{D}}} \big[ {\mathop {\max }\limits_{\left\| {{\bm{\varepsilon} _\mathbf{D}}} \right\| \le \rho } [{L_\mathbf{x}}({\mathbf{w}_t} + {\bm{\varepsilon} _\mathbf{D}}) - {L_\mathbf{x}}({\mathbf{w}_t})]} \big] \bigg|\\
&= \left| {\mathop {\rm{E}}\limits_{\mathbf{x} \in \mathbf{D}} \left[ {{\bm{\varepsilon} _\mathbf{D}}^{\rm T}{\nabla _\mathbf{w}}{L_\mathbf{x}}({\mathbf{w}_t})} \right] - \mathop {\rm{E}}\limits_{{\mathbf{x}} \in \hat{\mathbf{D}}} \left[ {{\bm{\varepsilon} _\mathbf{D}}^{\rm T}{\nabla _\mathbf{w}}{L_\mathbf{x}}({\mathbf{w}_t})} \right]} \right|\\
&= \frac{\rho }{{\left\| {{\nabla _\mathbf{w}}{L_\mathbf{D}}({\mathbf{w}_t})} \right\|}}\bigg| \mathop {\rm{E}}\limits_{\mathbf{x} \in {\mathbf{D}}} \left[ {{\nabla _\mathbf{w}}{L_\mathbf{D}}{{({\mathbf{w}_t})}^{\rm T}}{\nabla _\mathbf{w}}{L_\mathbf{x}}({\mathbf{w}_t})} \right] \\
&\quad\quad- \mathop {\rm{E}}\limits_{{\mathbf{x}} \in {\hat{\mathbf{D}}}} \left[ {{\nabla _\mathbf{w}}{L_\mathbf{D}}{{({\mathbf{w}_t})}^{\rm T}}{\nabla _\mathbf{w}}{L_\mathbf{x}}({\mathbf{w}_t})} \right] \bigg|\\
&= \frac{\rho }{{\left\| {{\nabla _\mathbf{w}}{L_\mathbf{D}}({w_t})} \right\|}}\bigg| \int_{\mathbf{x}} {{\nabla _\mathbf{w}}{L_\mathbf{D}}{{({\mathbf{w}_t})}^{\rm T}}{\nabla _\mathbf{w}}{L_\mathbf{x}}({\mathbf{w}_t})q ({\mathbf{x}})} d{\mathbf{x}} \\
&\quad\quad - \int_{\mathbf{x}} {p_\mathbf{x}^t{\nabla _\mathbf{w}}{L_\mathbf{D}}{{({\mathbf{w}_t})}^{\rm T}}{\nabla _\mathbf{w}}{L_\mathbf{x}}({\mathbf{w}_t})q ({\mathbf{x}})} d{\mathbf{x}} \bigg|\\
%&= \frac{\rho }{{\left\| {{\nabla _\mathbf{w}}{L_\mathbf{D}}({\mathbf{w}_t})} \right\|}}\bigg| \int_{\mathbf{x}} {\big[({\nabla _\mathbf{w}}{L_\mathbf{D}}{{({\mathbf{w}_t})}^{\rm T}}{\nabla _\mathbf{w}}{L_\mathbf{x}}({\mathbf{w}_t})) - p_\mathbf{x}^t({\nabla _\mathbf{w}}{L_\mathbf{D}}{{({\mathbf{w}_t})}^{\rm T}}{\nabla _\mathbf{w}}{L_\mathbf{x}}({\mathbf{w}_t})) \big]q ({\mathbf{x}})} d{\mathbf{x}} \bigg|\\
&\le \frac{\rho }{{\left\| {{\nabla _\mathbf{w}}{L_\mathbf{D}}({\mathbf{w}_t})} \right\|}}\int_{\mathbf{x}} \big| ({\nabla _\mathbf{w}}{L_\mathbf{D}}{{({\mathbf{w}_t})}^{\rm T}}({\nabla _\mathbf{w}}{L_\mathbf{x}}({\mathbf{w}_t}) - p_\mathbf{x}^t{\nabla _\mathbf{w}}{L_\mathbf{x}}({\mathbf{w}_t})) \big|q ({\mathbf{x}}) d{\mathbf{x}}\\
&\le \frac{\rho }{{\left\| {{\nabla _\mathbf{w}}{L_\mathbf{D}}({\mathbf{w}_t})} \right\|}}\int_{\mathbf{x}} {\left\| {{\nabla _\mathbf{w}}{L_\mathbf{D}}({\mathbf{w}_t})} \right\|\left\| {{\nabla _\mathbf{w}}{L_\mathbf{x}}({\mathbf{w}_t}) - p_\mathbf{x}^t{\nabla _\mathbf{w}}{L_\mathbf{x}}({\mathbf{w}_t})} \right\|q ({\mathbf{x}})} d{\mathbf{x}}\\
&= \rho \int_{\mathbf{x}} {\left\| {{\nabla _\mathbf{w}}{L_\mathbf{x}}({\mathbf{w}_t}) - p_\mathbf{x}^t{\nabla _\mathbf{w}}{L_\mathbf{x}}({\mathbf{w}_t})} \right\|q ({\mathbf{x}})} d{\mathbf{x}}\\
&= \rho \int_{\mathbf{x}} {(1 - p_\mathbf{x}^t)\left\| {{\nabla _\mathbf{w}}{L_\mathbf{x}}({\mathbf{w}_t})} \right\|q ({\mathbf{x}})} d{\mathbf{x}}.
\end{aligned}
\end{equation}
\end{small}
\end{proof}

\section{Hyperparameters settings}
For CIFAR-10 and CIFAR-100, our hyperparameter settings are consistent. The exact training hyperparameters on CIFAR-10/100 and Tiny-ImageNet are reported in Tab. \ref{subsec:hyp_cifar} and Tab. \ref{subsec:hyp_tiny}.
\begin{table*}[h]
\center
\caption{Hyperparameters settings for Cifar-10/100.}
\label{subsec:hyp_cifar}
%\resizebox{\linewidth}{!}{
\begin{tabular}{ccccccc}
\hline
Networks                          & Methods   & \begin{tabular}[c]{@{}c@{}}Peak learning\\ rate\end{tabular} & Weight decay            & $\rho$           & $s_{max}$          & $\alpha$         \\ \hline
\multirow{4}{*}{ResNet-18}        & SGD       & \multirow{4}{*}{0.05}                                        & \multirow{4}{*}{0.001}  & \textbackslash{} & \textbackslash{} & \textbackslash{} \\
                                  & SAM       &                                                              &                         & 0.1             & \textbackslash{} & \textbackslash{} \\
                                  & AUSAM-0.5 &                                                              &                         & 0.1             & 0.5              & 0.5              \\
                                  & AUSAM-0.6 &                                                              &                         & 0.1             & 0.5              & 0.6              \\ \hline
\multirow{4}{*}{WideResNet-28-10} & SGD       & \multirow{4}{*}{0.05}                                        & \multirow{4}{*}{0.001}  & \textbackslash{} & \textbackslash{} & \textbackslash{} \\
                                  & SAM       &                                                              &                         & 0.1              & \textbackslash{} & \textbackslash{} \\
                                  & AUSAM-0.5 &                                                              &                         & 0.1              & 0.5              & 0.5              \\
                                  & AUSAM-0.6 &                                                              &                         & 0.1              & 0.5              & 0.6              \\ \hline
\multirow{4}{*}{PyramidNet-110}   & SGD       & \multirow{4}{*}{0.1}                                         & \multirow{4}{*}{0.0005} & \textbackslash{} & \textbackslash{} & \textbackslash{} \\
                                  & SAM       &                                                              &                         & 0.2              & \textbackslash{} & \textbackslash{} \\
                                  & AUSAM-0.5 &                                                              &                         & 0.2              & 1                & 0.5              \\
                                  & AUSAM-0.6 &                                                              &                         & 0.2              & 1                & 0.6              \\ \hline
\end{tabular}%}
\end{table*}

\begin{table*}[h]
\center
\caption{Hyperparameters settings for Tiny-ImageNet.}
\label{subsec:hyp_tiny}
%\resizebox{\linewidth}{!}{
\begin{tabular}{ccccccc}
\hline
Networks                     & Methods   & \begin{tabular}[c]{@{}c@{}}Peak learning\\ rate\end{tabular} & Weight decay            & $\rho$           & $s_{max}$        & $\alpha$         \\ \hline
\multirow{4}{*}{ResNet-18}   & SGD       & \multirow{4}{*}{0.05}                                        & \multirow{4}{*}{0.001}  & \textbackslash{} & \textbackslash{} & \textbackslash{} \\
                             & SAM       &                                                              &                         & 0.1             & \textbackslash{} & \textbackslash{} \\
                             & AUSAM-0.5 &                                                              &                         & 0.1             & 0.5              & 0.5              \\
                             & AUSAM-0.6 &                                                              &                         & 0.1             & 0.5              & 0.6              \\ \hline
\multirow{4}{*}{ResNet-50}   & SGD       & \multirow{4}{*}{0.1}                                         & \multirow{4}{*}{0.0005} & \textbackslash{} & \textbackslash{} & \textbackslash{} \\
                             & SAM       &                                                              &                         & 0.1              & \textbackslash{} & \textbackslash{} \\
                             & AUSAM-0.5 &                                                              &                         & 0.1              & 1                & 0.5              \\
                             & AUSAM-0.6 &                                                              &                         & 0.1              & 1                & 0.6              \\ \hline
\multirow{4}{*}{MobileNetv1} & SGD       & \multirow{4}{*}{0.05}                                        & \multirow{4}{*}{0.0005} & \textbackslash{} & \textbackslash{} & \textbackslash{} \\
                             & SAM       &                                                              &                         & 0.05             & \textbackslash{} & \textbackslash{} \\
                             & AUSAM-0.5 &                                                              &                         & 0.05             & 1                & 0.5              \\
                             & AUSAM-0.6 &                                                              &                         & 0.05             & 1                & 0.6              \\ \hline
\end{tabular}%}
\end{table*}
When applying SAM and AUSAM to SimCC and QAT, we maintain most of the hyperparameter settings of SimCC and QAT, respectively. The partially changed hyperparameters are shown in Tab. \ref{subsec:hyp_simcc} and Tab. \ref{subsec:hyp_qat}, respectively.
\begin{table*}[!h]
\center
\caption{Hyperparameters settings for SimCC.}
\label{subsec:hyp_simcc}
%\resizebox{\linewidth}{!}{
\begin{tabular}{ccccccc}
\hline
Methods    & \begin{tabular}[c]{@{}c@{}}Batch\\ size\end{tabular} & Image size               & \begin{tabular}[c]{@{}c@{}}learning\\ rate\end{tabular} & \begin{tabular}[c]{@{}c@{}}Weight\\ decay\end{tabular} & $\rho$           & $s_{max}$        \\ \hline
Adam       & \multirow{3}{*}{64}                                  & \multirow{3}{*}{256$\times$256} & \multirow{3}{*}{0.001}                                  & \multirow{3}{*}{0.0001}                                & \textbackslash{} & \textbackslash{} \\
SAM+Adam   &                                                      &                          &                                                         &                                                        & 0.05             & \textbackslash{} \\
AUSAM+Adam &                                                      &                          &                                                         &                                                        & 0.05             & 1                \\ \hline
\end{tabular}%}
\end{table*}

\begin{table*}[!h]
\center
\caption{Hyperparameters settings for QAT.}
%\resizebox{\linewidth}{!}{
\label{subsec:hyp_qat}
\begin{tabular}{ccccccc}
\hline
Networks                     & Methods   & Batch size           & Learning rate          & Weight decay            & $\rho$           & $s_{max}$        \\ \hline
\multirow{3}{*}{ResNet-18}   & QAT+SGD   & \multirow{3}{*}{128} & \multirow{3}{*}{0.005} & \multirow{3}{*}{0.0005} & \textbackslash{} & \textbackslash{} \\
                             & QATSAM    &                      &                        &                         & 0.05             & \textbackslash{} \\
                             & QAT+AUSAM &                      &                        &                         & 0.05             & 0.5              \\ \hline
\multirow{3}{*}{MobileNetv1} & QAT+SGD   & \multirow{3}{*}{128} & \multirow{3}{*}{0.005} & \multirow{3}{*}{0.0005} & \textbackslash{} & \textbackslash{} \\
                             & QATSAM    &                      &                        &                         & 0.05             & \textbackslash{} \\
                             & QAT+AUSAM &                      &                        &                         & 0.05             & 0.5              \\ \hline
\end{tabular}%}
\end{table*}